\documentclass[11pt,a4paper]{article}
\usepackage[hyperref]{acl2020}
\usepackage{tikz-dependency}
\usepackage[title]{appendix}
\usepackage{nicefrac}
\usepackage{siunitx}
\usepackage{multirow}
\usepackage{array,booktabs}
\usepackage{times}
\usepackage{amsmath}
\usepackage{dcolumn}
\usepackage{contour}
\usepackage[normalem]{ulem}

\aclfinalcopy 
\def\aclpaperid{xxx} 

\contourlength{0.8pt}

\newcommand{\myuline}[1]{%
  \uline{\phantom{#1}}%
  \llap{\contour{white}{#1}}%
}

\newcolumntype{d}[1]{D{.}{.}{#1}}

\title{Efficient EUD Parsing}

\author{Mathieu Dehouck \hfill  Mark Anderson \hfill Carlos G\'{o}mez-Rodr\'{i}guez\\
  Universidade da Coru\~na, CITIC \\
  FASTPARSE Lab, LyS Research Group, \\ 
  Departamento de Ciencias de la Computaci\'{o}n y Tecnolog\'{i}as de la Informaci\'{o}n \\
  Campus Elvi\~{n}a, s/n, 15071 
  A Coru\~{n}a, Spain\\
  {\texttt \{mathieu.dehouck,m.anderson,carlos.gomez\}@udc.es}}

\date{}

\definecolor{deppink}{HTML}{a92b7a}
\definecolor{depgreen}{HTML}{679033}
\definecolor{deporange}{HTML}{be6320}
\definecolor{depblue}{HTML}{165f77}
\definecolor{depred}{HTML}{b12019}

\newcommand{\john}[1]{\textcolor{black}{#1}}
\definecolor{LightMagenta}{RGB}{249, 249, 255}
\begin{document}

\maketitle
\begin{abstract}
We present the system submission from the FASTPARSE team for the EUD Shared Task at IWPT 2020. We engaged with the task by focusing on efficiency. For this we considered training costs and inference efficiency. Our models are a combination of distilled neural dependency parsers and a rule-based system that projects UD trees into EUD graphs. We obtained an average ELAS of 74.04 for our official submission, ranking 4th overall.  
\end{abstract}

\section{Introduction}
Latterly, the environmental impact of AI and NLP's dependency on deep neural networks has come under scrutiny \citep{schwartz2019green,strubell2019energy}. This has coincided with a renewed push for efficiency in NLP so as to make systems more easily used in different contexts, be it in hardware impaired conditions, large web-scale applications, or a host of other considerations \citep{strzyz2019viable,clark2019bam,vilares2019better,junczys2018marian}. 

Here we describe our contribution to the Enhanced Universal Dependencies (EUD) Shared Task at IWPT 2020 \citep{EUD2020}, where we have considered efficiency as well as bare accuracy performance. We combine linguistics and machine learning to develop efficient parsers, both with respect to training and inference. First we curtail the amount of training data we use, second we try distillation to create smaller networks for dependency parsers while maintaining accuracy, and third we develop a rule-based system to cast universal dependency (UD) trees as EUD graphs.
\subsection{\john{An aside on enhanced graphs}}
\john{Certain syntactic phenomena, such as the propagation of conjuncts or coreferences in relative clauses, can only be handled implicitly by Universal Dependency (UD) trees resulting in opaque relations or long paths between related content words. EUD graphs is an enhanced representation which can handle these phenomena explicitly. As nodes are not restricted to a single head, these more complex relations can be more readily represented. While this results in a potentially much more useful and informative representation, it also makes for a more challenging task than vanilla UD parsing.}  
\section{Forest felling}\label{sec:felling_analysis}
Distillation introduces extra training overheads. To mitigate this and to balance our pursuit of inference efficiency with some semblance of training efficiency and considering recent results using distillation suggest larger treebanks suffer greater \citep{anderson2020}, 
we decided to set a limit to the size of training treebanks.  

In order to minimise introducing compounding variables that could affect training efficacy, we renormalise the sampled treebanks to follow the same tree length distribution of the original treebank. Where more than one treebank exists for a given language, we took a sample from each treebank renormalised with respect to that treebank and took a sample size so that the contribution from each treebank would follow the same ratio as the full data for that language.

We evaluated what limit to set by testing on 4 languages spanning 3 language families (Uralic, Afro-Asiatic, and Indo-European). The only family to appear in the shared task training data not covered was Dravidian as the only example from this language, Tamil, has too small a treebank to have been useful for this analysis. We also cover two branches of the Indo-European family. 
Balto-Slavic is covered by Russian as the treebank is rather large and uses the Cyrillic script. Germanic is covered by Dutch, which we chose as there are two treebanks which combine to a sizeable number of trees and so would cover the case of combining different treebanks. Finnish was used to cover the Uralic family as we carried this experiment out before the larger Estonian treebank was made available and Arabic was used for Afro-Asiatic. 
We used sample treebank sizes of 1,000, 3,000, 6,075 (the number of trees in the Arabic treebank), 12,217 (the number of trees in the Finnish treebank), and 18,051 (the combined number of trees in both Dutch treebanks). We created 2 splits where possible (i.e. at 6,075 trees Arabic isn't a sample treebank) as a limited attempt at experimental robustness.
\begin{figure}[tbp]
    \centering
    \includegraphics[width=0.99\linewidth]{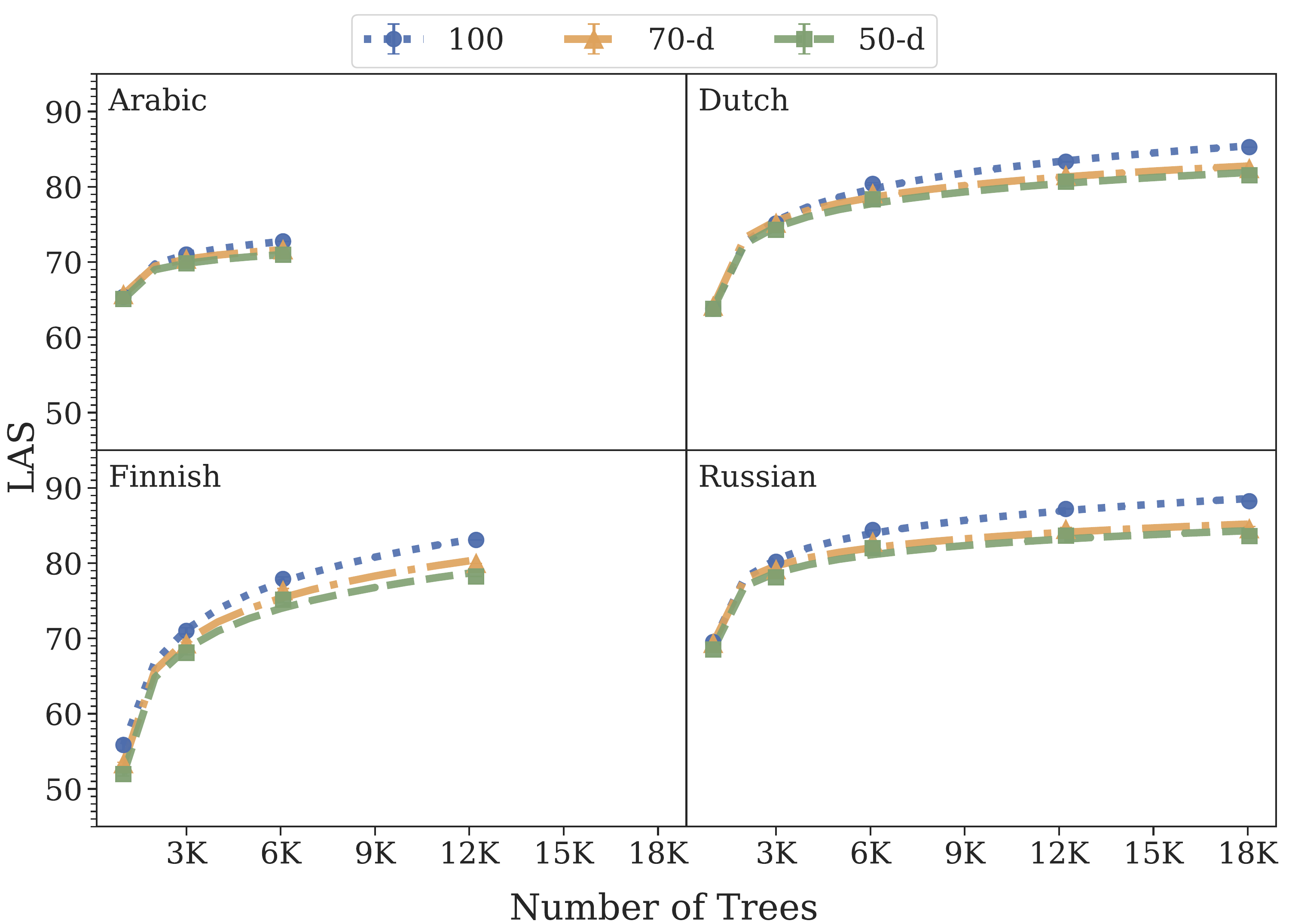}
    \caption{LAS for different models for Arabic, Dutch, Finnish, and Russian development treebanks.}
    \label{fig:las}
\end{figure}

We train a Biaffine parser using the hyperparameters of the original paper, shown in Table \ref{tab:experimental_hyperparameters} in the Appendix \cite{dozat20161}. We then distill (as described in Section \ref{sec:distill} and in Appendix \ref{sec:teacher}) these models to two different network sizes, one which has 70\% of the number of nodes in both the BILSTM and MLP layers and one that has 50\%. Otherwise the structure of the network is the same as the base model. The LAS averaged over the splits for each sample and model are shown in Figure \ref{fig:las} (similarly for UAS in Figure \ref{fig:uas} in Appendix \ref{sec:teacher}). We are limited by what we can extrapolate from the results for Arabic and Finnish other than they appear to follow a similar trend to Dutch and Russian. For the latter languages we observe the performance levelling at larger treebank sizes, which is neither remarkable nor unexpected, but also a widening between the performance of the full and the distilled models. 
\begin{figure}[tbp]
    \centering
    \includegraphics[width=0.99\linewidth]{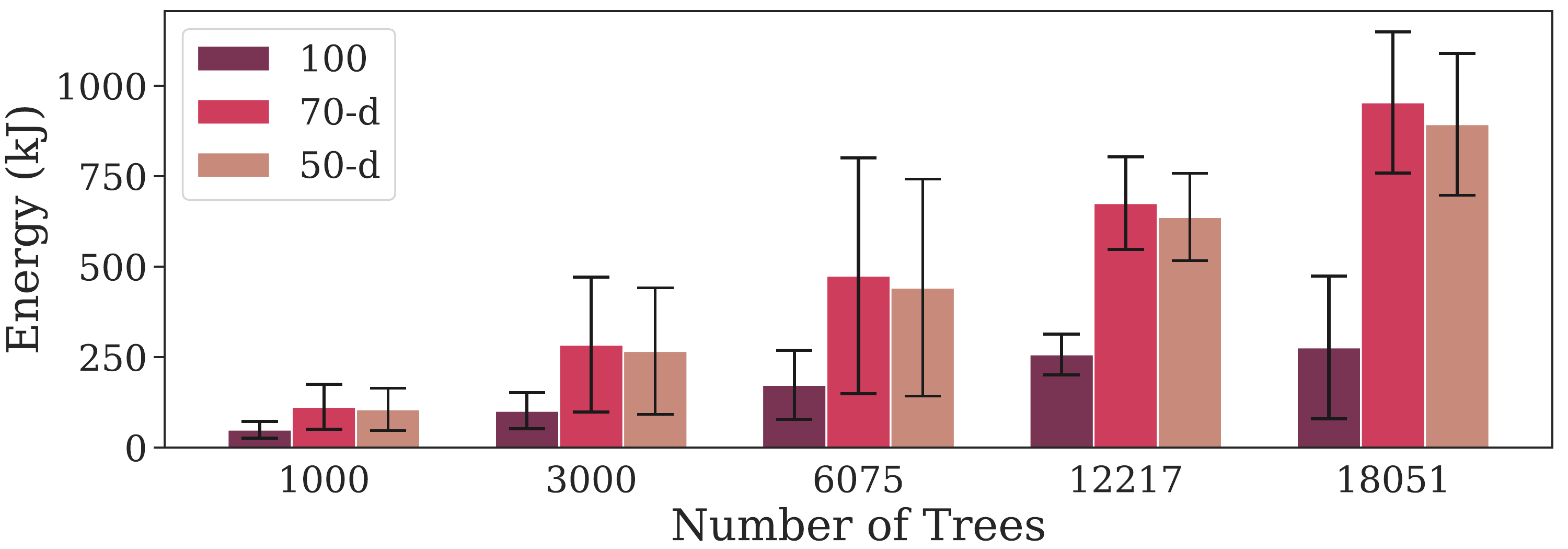}
    \caption{Training energy consumption for different models for different treebank sizes averaged over Arabic, Dutch, Finnish, and Russian.}
    \label{fig:power}
\end{figure}

As we are concerned with training efficiency, we present the energy consumption for each model type averaged over language and split in Figure \ref{fig:power}. The amount of energy required to distill our models increases significantly with respect to treebank size. However, distilling to a smaller model requires less energy and, as can be seen in Figure \ref{fig:las}, the accuracy difference between the two distilled models is nominal. 
\begin{figure}[tbp]
    \centering
    \includegraphics[width=0.99\linewidth]{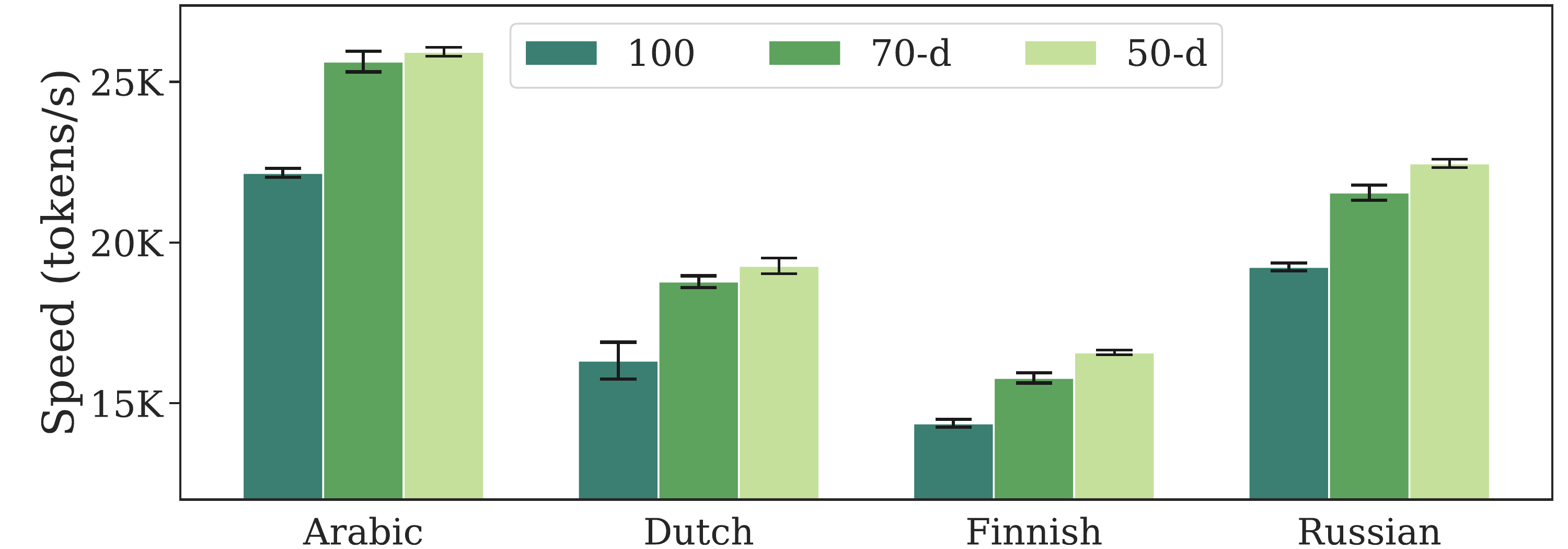}
    \caption{GPU inference speed for different models for treebank 12k (except Arabic which uses its full treebank of 6075 trees) averaged over 5 runs on the development treebanks with batch size 256.}
    \label{fig:speed}
\end{figure}

Figure \ref{fig:speed} shows the inference speed (averaged over splits and 5 runs) on GPU using a single CPU core for each language using the models trained with the 12,217 treebanks (for Arabic we use its full treebank). We observe a sizeable increase in speed over the baseline model for both distilled models, but only a small difference between the two distilled models.

From this, we decided to set an upper limit on the treebank size for the main task to 13,121 (the size of the Italian treebank) as this would require the least amount of tampering and was close to the 
second largest treebank size
used here which performed close to the largest. This meant taking a sample of the Czech, Dutch, Estonian, and Polish treebanks and combining them as described above. A sample was taken for the Russian treebank. Some syntactic metrics are given in Table \ref{tab:treebank_samples} in the Appendix which shows the different breakdown of the training data used for each of these languages and how they are very similar to the full data. Also, we opted to distill to 50\% of the original model size. 
For this analysis, and all subsequent analyses, the CPU used was an Intel Core i7-7700 and the GPU an Nvidia GeForce GTX 1080.\footnote{Using Python 3.7.0, PyTorch 1.0.0, and CUDA 8.0.}

\section{Boiling neural networks in the pot still}\label{sec:distill}
Neural network compression is not a new phenomenon. For example, pruning has long been shown to be an effective way to reduce parameters with minimal impact on accuracy and also to help generalisation \cite{lecun1990,hagiwara1994,wan2009,han2015learning,see2016}. However, pruning isn't overly useful for creating efficient models as they leave networks in irregularly sparse states. Other techniques exist that can recast networks into smaller more efficient ones, but we focus on distillation. For a detailed survey of current neural network compression techniques see \citet{cheng2017survey}. 

\citet{anderson2020} used \textit{teacher-student} distillation to increase the inference efficiency while only losing marginal accuracy for Universal Dependency (UD) parsing, showing that distilled models outperform models of the same structure and size trained normally. Here we extend that work and use \textit{teacher-student} distillation to obtain efficient dependency parsers as the basis of our enhanched-dependency parser systems. A full description of our implementation can be found in \citet{anderson2020} but we also offer a condensed version in Appendix \ref{sec:teacher}.


While we curtailed our training data, we selected our models based on the performance on the full development data for a given language with gold sentence segmentation and tokenisation. 
We used characters and words as input to our network. The embeddings for both were randomly initialised. The hyperparameters are the same as used above. 
We also used early stopping to limit unnecessary training time, stopping after 10 epochs without performance improvement.

At inference time we used UDPipe v2.5 models to predict everything except the parse \cite{straka2020}. When a combination of treebanks were being predicted, we used the model which corresponded to the largest of the treebanks.

 
\begin{table}
    \centering
    \small
    \begin{tabular}{lcd{5.1}}
    \toprule
     \multicolumn{3}{c}{Training costs} \\
    \midrule
    & Total time & \multicolumn{1}{c}{GPU Energy (kJ)}\\
     \textbf{Base} & 08h:42m:52.1s & 3570.7 \\
\textbf{Distill} & 30h:07m:49.6s & 9981.8 \\
     \textbf{Rule-based} & 00h:00m:41.1s & n/a \\
     \bottomrule
    \end{tabular}
    \caption{Total training time and GPU energy consumption for all treebanks.} 
    \label{tab:training_costs}
\end{table}
Table \ref{tab:training_costs} shows the total time to train the full-sized models and the distillation models for all languages. Also, shown is the GPU energy consumption. 
The costs for distillation include those of the base models.

Training costs for distillation are more than three times that of the baseline which is hardly surprising. The inference energy cost for all development treebanks (37K trees) for the full model is 2.10 (0.09)kJ (average value over 5 runs for each treebank) whereas the cost for distillation is 1.49 (0.03)kJ. Based on these measurements, we would need to parse 390M sentences to offset the extra cost of distilling models when running on GPU. 

\begin{table}[b]
\centering
\tabcolsep=.049cm
\small
\begin{tabular}{rccccrccc}
\toprule
&\multicolumn{1}{c}{UAS} & LAS & ELAS & \phantom{} & & \multicolumn{1}{c}{UAS} & LAS   & ELAS   \\\midrule
\multirow{1}{*}{\textit{\textbf{Arabic}}} & & & & & \multirow{1}{*}{\textit{\textbf{Bulgarian}}} \\full & \multicolumn{1}{c}{77.0} & 72.8 & 68.4 &  & full& \multicolumn{1}{c}{91.5} & 87.6 & 85.3 \\dist & \multicolumn{1}{c}{76.5} & 72.3 & 67.9 &  & dist& \multicolumn{1}{c}{91.6} & 87.6 & 85.2 \\udpipe & \multicolumn{1}{c}{72.8} & 68.1 & 63.0 &  & udpipe& \multicolumn{1}{c}{88.7} & 84.3 & 81.9 \\
\multirow{1}{*}{\textit{\textbf{Czech}}} & & & & & \multirow{1}{*}{\textit{\textbf{Dutch}}} \\full & \multicolumn{1}{c}{90.0} & 87.0 & 82.4 &  & full& \multicolumn{1}{c}{87.5} & 84.0 & 82.2 \\dist & \multicolumn{1}{c}{89.0} & 85.3 & 80.7 &  & dist& \multicolumn{1}{c}{86.7} & 82.9 & 81.0 \\udpipe & \multicolumn{1}{c}{87.6} & 84.0 & 78.4 &  & udpipe& \multicolumn{1}{c}{79.3} & 75.0 & 73.2 \\
\multirow{1}{*}{\textit{\textbf{English}}} & & & & & \multirow{1}{*}{\textit{\textbf{Estonian}}} \\full & \multicolumn{1}{c}{85.6} & 82.6 & 81.2 &  & full& \multicolumn{1}{c}{85.5} & 81.5 & 80.3 \\dist & \multicolumn{1}{c}{84.4} & 81.2 & 79.8 &  & dist& \multicolumn{1}{c}{84.7} & 80.2 & 79.0 \\udpipe & \multicolumn{1}{c}{81.0} & 77.6 & 76.3 &  & udpipe& \multicolumn{1}{c}{81.5} & 77.6 & 76.7 \\
\multirow{1}{*}{\textit{\textbf{Finnish}}} & & & & & \multirow{1}{*}{\textit{\textbf{French}}} \\full & \multicolumn{1}{c}{86.2} & 83.1 & 79.9 &  & full& \multicolumn{1}{c}{88.1} & 85.5 & 82.3 \\dist & \multicolumn{1}{c}{85.1} & 81.3 & 78.0 &  & dist& \multicolumn{1}{c}{88.5} & 85.8 & 82.6 \\udpipe & \multicolumn{1}{c}{80.4} & 76.8 & 73.7 &  & udpipe& \multicolumn{1}{c}{85.2} & 82.6 & 79.4 \\
\multirow{1}{*}{\textit{\textbf{Italian}}} & & & & & \multirow{1}{*}{\textit{\textbf{Latvian}}} \\full & \multicolumn{1}{c}{91.6} & 89.3 & 87.8 &  & full& \multicolumn{1}{c}{86.7} & 83.2 & 79.3 \\dist & \multicolumn{1}{c}{90.3} & 87.8 & 85.9 &  & dist& \multicolumn{1}{c}{86.0} & 81.9 & 78.2 \\udpipe & \multicolumn{1}{c}{88.5} & 85.9 & 84.1 &  & udpipe& \multicolumn{1}{c}{79.8} & 75.4 & 70.5 \\
\multirow{1}{*}{\textit{\textbf{Lithuanian}}} & & & & & \multirow{1}{*}{\textit{\textbf{Polish}}} \\full & \multicolumn{1}{c}{77.6} & 72.7 & 68.6 &  & full& \multicolumn{1}{c}{90.9} & 87.2 & 78.6 \\dist & \multicolumn{1}{c}{78.0} & 73.0 & 68.9 &  & dist& \multicolumn{1}{c}{90.2} & 86.0 & 77.2 \\udpipe & \multicolumn{1}{c}{72.3} & 64.6 & 60.9 &  & udpipe& \multicolumn{1}{c}{87.1} & 82.6 & 74.7 \\
\multirow{1}{*}{\textit{\textbf{Russian}}} & & & & & \multirow{1}{*}{\textit{\textbf{Slovak}}} \\full & \multicolumn{1}{c}{90.2} & 87.3 & 84.4 &  & full& \multicolumn{1}{c}{85.4} & 81.6 & 77.0 \\dist & \multicolumn{1}{c}{88.9} & 85.5 & 82.5 &  & dist& \multicolumn{1}{c}{84.7} & 80.7 & 76.1 \\udpipe & \multicolumn{1}{c}{87.4} & 84.4 & 81.5 &  & udpipe& \multicolumn{1}{c}{81.2} & 75.9 & 70.5 \\
\multirow{1}{*}{\textit{\textbf{Swedish}}} & & & & & \multirow{1}{*}{\textit{\textbf{Tamil}}} \\full & \multicolumn{1}{c}{85.2} & 81.4 & 78.9 &  & full& \multicolumn{1}{c}{59.8} & 52.6 & 51.2 \\dist & \multicolumn{1}{c}{85.3} & 81.6 & 79.0 &  & dist& \multicolumn{1}{c}{64.0} & 56.9 & 55.5 \\udpipe & \multicolumn{1}{c}{79.5} & 75.4 & 73.2 &  & udpipe& \multicolumn{1}{c}{60.7} & 54.1 & 53.0 \\
\multirow{1}{*}{\textit{\textbf{Ukrainian}}} & & & & & \multirow{1}{*}{\textit{\textbf{Average}}} \\full & \multicolumn{1}{c}{87.1} & 83.2 & 78.3 &  & full& \multicolumn{1}{c}{85.0} & 81.3 & 78.0 \\dist & \multicolumn{1}{c}{86.6} & 82.5 & 77.5 &  & dist& \multicolumn{1}{c}{84.7} & 80.7 & 77.3 \\udpipe & \multicolumn{1}{c}{81.6} & 76.9 & 72.5 &  & udpipe& \multicolumn{1}{c}{80.9} & 76.5 & 73.1 \\\bottomrule
\end{tabular}
\caption{Attachment scores for both UD trees and EUD graphs for the development treebanks using different dependency parsers: full baseline models (Full), distilled models (dist), and UDPipe v2.5 models (udpipe).}
\label{tab:dev_performance}
\end{table}

Late in the day we decided to validate the results of \citet{anderson2020}, namely that distilled models outperform models trained normally of equivalent sizes. This highlighted that our distilled models used for our official score had not converged. We trained new distilled models and the results given here are for these new models. Our official results using the partially-trained models are in table \ref{tab:test_results_full_old} in the Appendix. All results, including training costs, in this section are for the full-trained distilled models and unless otherwise stated are using the combined development treebanks for each language. 

Table \ref{tab:small_performance} in the Appendix shows the performance for the equivalent-sized models trained normally (small) and the distilled models (dist) with respect to UAS and LAS. For the most part the normal models outperform the distilled models. The main differences between our work and that of \citet{anderson2020} is we do not use pre-trained word embeddings nor POS tags as features. 
So perhaps without this extra information distillation is less effective. Also, dropout wasn't used during distillation in the original paper but is here, so perhaps the values used here were too punitive a regularisation. Although we use the same hyperparameters as the original paper, the average LAS for the small normally trained models is 0.4 points less than the large model.  

We also evaluated the distilled models against the full baseline model and UDPipe v2.5. 
These results are shown in Table \ref{tab:dev_performance}. The distilled models outperform the UDPipe models and are within a point of both UAS and LAS to the full model. The ELAS results for the rule-based system using the predicted dependency trees from each of theses systems are also shown. 
The performance on ELAS generally follows the dependency scores. 

\begin{figure}[htbp!]
    \centering
    \includegraphics[width=0.99\linewidth]{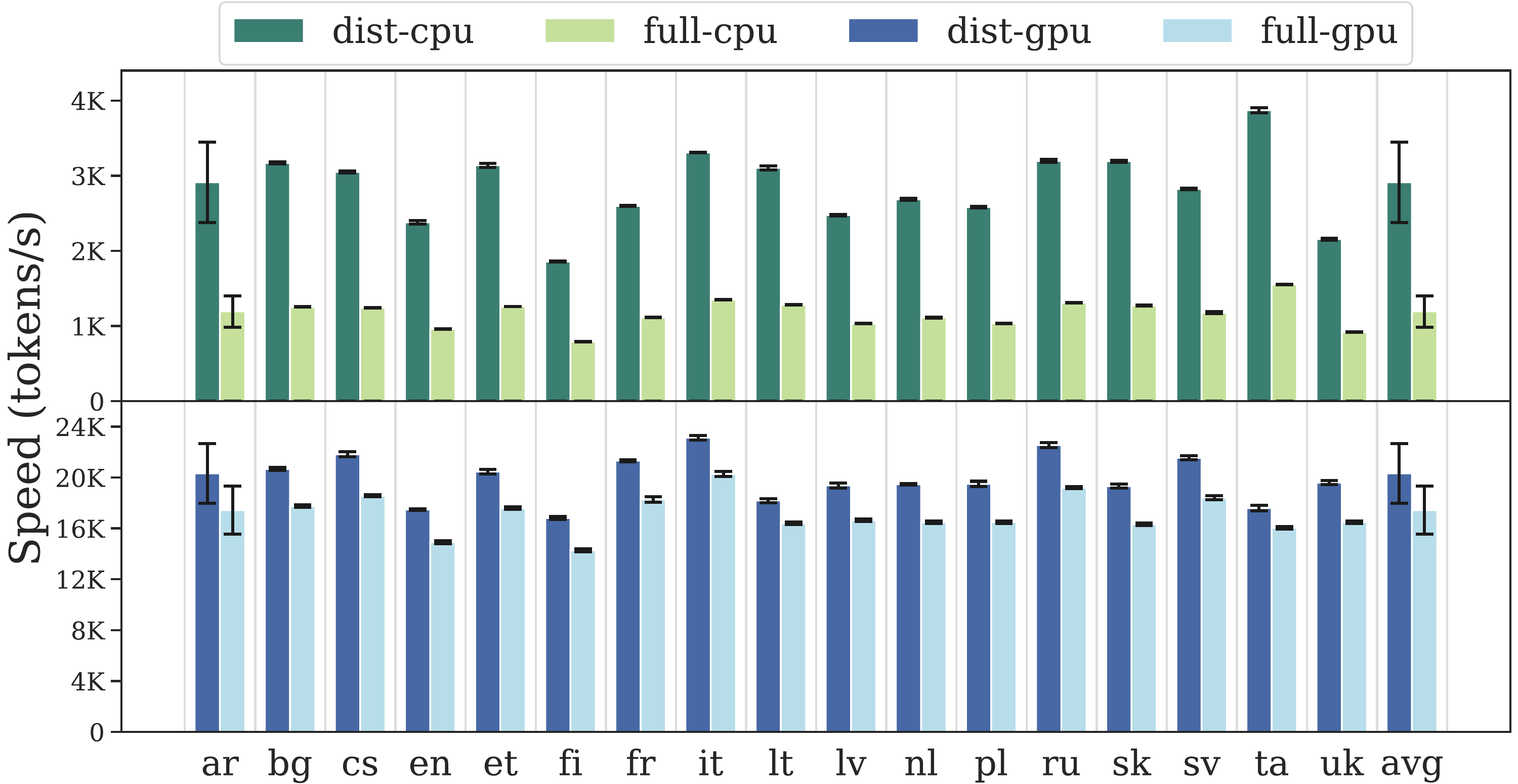}
    \caption{Inference speed for distilled (dist) and full baseline models on CPU (-cpu) and GPU (-gpu) for each development treebank averaged over 5 runs using one CPU core with batch size 256.}
    \label{fig:dev_speed}
\end{figure}
\begin{table}
    \newcolumntype{C}{ @{}>{${}}c<{{}$}@{} }
    \tabcolsep=.02cm
    \centering
    \small
    \begin{tabular}{>{\raggedright\arraybackslash}p{2.5em}>{\raggedright\arraybackslash}p{4em} rl p{1em} rl}
    \toprule
    \multicolumn{7}{c}{Inference speed (token/s)} \\\midrule
    
   \multirow{2}{*}{CPU} && \multicolumn{2}{c}{UD parser} & &\multicolumn{2}{c}{Full pipeline}\\

    &\textbf{Base} & 1194.1 &(207.1)& & 879.0 & (123.4)\\
     & \textbf{Distill} & 2912.9 &(535.1)& & 1569.9 &(238.8) \\
    & \textbf{UDPipe} & 3629.4   &(584.0)& &2220.2&(698.0) \\
    \multirow{2}{*}{GPU} \\
    & \textbf{Base} &  17427.0& (1890.3)& &  2993.3 & (680.2)\\
     & \textbf{Distill} & 20321.6 & (2348.9) & & 3073.7 & (714.9) \\
     \bottomrule
    \end{tabular}
    \caption{Inference speeds for dependency parsers and the full EUD pipeline for different systems run on development treebanks and averaged over 5 runs.}
    \label{tab:speeds}
\end{table}
Figure \ref{fig:dev_speed} shows the inference speed using GPU and CPU of the full baseline model and the distilled models for each language. These are obtained by running the parser 5 times for each language on the full development data and only using one CPU core. 
The average speed (token/second) increase was 2.44x (1.17x) on CPU (GPU). 

Table \ref{tab:speeds} shows the inference speeds for the full pipeline and the dependency parser. 
We also compare UDPipe inference performance as it is a viable candidate for an efficient parser. 
It is the fastest of the systems compared here, but the full pipeline which used it obtained an average ELAS 4.9 points less than full baseline model whereas the distilled models are only 0.7 points less. 

\section{Unravelling trees with shrewd rules}

Rule-based systems are intrinsically efficient with respect to training time (barely a flash in the pan) and inference time (there is practically none). So we developed a simple rule-based system to enhance the existing dependency tree and reveal hidden dependencies in a cross-lingual setting using 
as few language specific rules as possible.
Beyond the basic enhancement of the original dependencies, there are four main phenomena that create new dependencies: relative clauses, control, conjunction and ellipsis.
Since our pipeline does not predict empty nodes, we decided to ignore ellipsis in this system. To deal with each of these phenomena, our algorithm needs to make a number of passes over each sentence. 

\paragraph{Pass one - relative clauses and controls:}
The first pass of the algorithm iterates through each word in the sentence and creates enhanced relations according to the type of the original dependency.
When necessary, it adds lemma and case information. If the current word is a relative pronoun/adverb, its antecedent is found by following its path to the root until an \texttt{acl:relcl} relation is met.
Then a \texttt{ref} edge is created between the word and its antecedent and an edge between the antecedent and the governor of the relativiser with the same relation type as the original relation (if the relative pronoun is the object of a verb then the antecedent becomes the object of that verb).
If the word is the dependent of an \texttt{xcomp} relation, the algorithm looks for a subject amongst its controlling predicate's arguments.
If a subject is found, it creates an edge between the subject and the current word of type \texttt{nsubj(:xsubj)} (or \texttt{csubj} in the case of a clausal argument).
If no subject is available, the current word is stored in a separate list for later processing.
If the word is the dependent of a \texttt{conj} relation, it too is stored in a separate list 
along with all other conjuncts. 
Whenever we encounter an argument of the type subject, object or oblique, this information is kept for resolving subjects of controlled predicates.
\begin{figure}[htpb!]
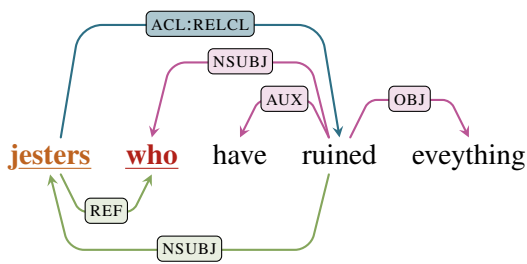

\centering
\begin{dependency}[edge style={deppink!80, thick},label style={fill=deppink!15},edge slant=7]
\begin{deptext}[column sep=0.75em,ampersand replacement=\^]
\textcolor{deporange}{\textbf{{\myuline{jesters}}}} \^ \textcolor{depred}{\textbf{{\myuline{who}}}} \^ have \^ ruined \^ eveything \\
\end{deptext}
\depedge{4}{2}{\textsc{nsubj}}
\depedge{4}{3}{\textsc{aux}}
\depedge[edge style={depblue!90, thick},label style={fill=depblue!35}]{1}{4}{\textsc{acl:relcl}}
\depedge{4}{5}{\textsc{obj}}
\depedge[edge below,edge style={depgreen!80, thick},label style={fill=depgreen!15}]{1}{2}{\textsc{ref}}
\depedge[edge below,edge height=1cm, edge style={depgreen!80, thick},label style={fill=depgreen!15}]{4}{1}{\textsc{nsubj}}
\end{dependency}
\caption{Relative clause example. Pre-existing edges in graph are in magenta and blue. The algorithm observes an \texttt{acl:relcl} relation (highlighted in blue) which causes it to generate two new relations (highlighted in green). A \texttt{ref} relation is created between \textcolor{depred}{\textbf{{who}}} and its antecedent, \textcolor{deporange}{\textbf{{jesters}}}. Then a \texttt{nsubj} is propagated from the head of \textcolor{depred}{\textbf{{who}}}, \textit{ruined}, to \textcolor{deporange}{\textbf{{jesters}}}.}\label{fig:rel}
\end{figure}
\paragraph{Pass two - resolving conjunctions:}
We have two general functions, one for dependent level conjunctions and one for governor level conjunctions, and a few special cases.
The dependent level function propagates the conjunction head's original relation to its conjuncts adapting it if necessary, for example in coordinated \texttt{nmod} with different adposition or case. 
The governor level function propagates the conjunction head's dependents to its conjuncts in the absence of similar dependents and according to morphological agreement. 
We have a special function that handles subjects of conjuncts because subjects are more diverse than other syntactic functions.
In UD at least three relations can mark subjects, namely \texttt{nsubj} for nominal subjects, \texttt{csubj} for clausal subjects and \texttt{expl} used amongst other for syntactic subjects in non prodrop languages (e.g. \textit{"it rains"}). 
Subject edges 
also embed information about their governor, notably information about the voice as \texttt{:pass} when relevant.
And, subjects can be absent altogether in prodrop languages, so we rely on morphological information to decide to propagate a given subject in these languages.

\begin{figure}[htpb!]
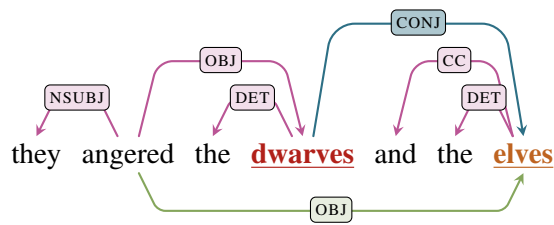

\centering
\begin{dependency}[edge style={deppink!80, thick},label style={fill=deppink!15},edge slant=7]
\begin{deptext}[column sep=0.25em,ampersand replacement=\^]
they \^ angered \^ the \^ \textcolor{depred}{\textbf{\underline{{dwarves}}}} \^ and \^ the \^ \textcolor{deporange}{\textbf{\underline{{elves}}}} \\  
\end{deptext}
\depedge{2}{1}{\textsc{nsubj}}
\depedge{4}{3}{\textsc{det}}
\depedge{2}{4}{\textsc{obj}}
\depedge{7}{5}{\textsc{cc}}
\depedge{7}{6}{\textsc{det}}
\depedge[edge style={depblue!90, thick},label style={fill=depblue!35}]{4}{7}{\textsc{conj}}
\depedge[edge below,edge height=0.5cm,edge style={depgreen!80, thick},label style={fill=depgreen!15}]{2}{7}{\textsc{obj}}
\end{dependency}
\caption{Conjunction example.  Magenta and blue edges are those existing in the graph after one pass. During the first pass \textcolor{deporange}{\textbf{{elves}}} is stored as it is the dependent of a \texttt{conj} relation (highlighted in blue). On the second pass the \texttt{obj} relation of \textcolor{depred}{\textbf{{dwarves}}}, the head of this \texttt{conj} relation, propagates to \textcolor{deporange}{\textbf{{elves}}} generating a new \texttt{obj} relation (highlighted in green) from \textit{angered}.}\label{fig:conj}
\end{figure}

\paragraph{Pass three and onwards - sweeping up controls:}
Once conjunctions have been resolved and more predicates have their arguments stored, the algorithm iterates over controlled predicates that do not have a subject after the first sentence traversal.
 Several such iterations may be necessary since the number of times a predicate may be coordinated with a controlled verb itself already coordinated to another controlled verb is not bounded.
Like in the sentence \textit{``Sam stood up and wanted to scream and start running.''}
But in practice one iteration solves the vast majority of missing subjects.

\begin{figure}[htpb!]
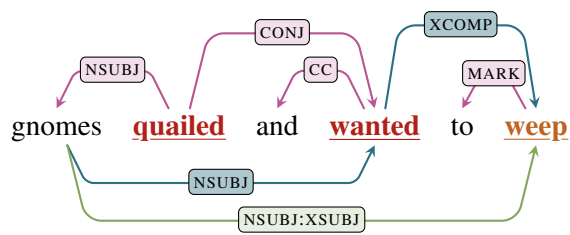

\centering
\begin{dependency}[edge style={deppink!80, thick},label style={fill=deppink!15},edge slant=7]
\begin{deptext}[column sep=0.6em,ampersand replacement=\^]
gnomes \^ \textcolor{depred}{\textbf{{\myuline{quailed}}}} \^ and  \^ \textcolor{depred}{\textbf{{\underline{wanted}}}} \^ to \^ \textcolor{deporange}{\textbf{{\myuline{weep}}}} \\
\end{deptext}
\depedge{2}{1}{\textsc{nsubj}}
\depedge{4}{3}{\textsc{cc}}
\depedge{2}{4}{\textsc{conj}}
\depedge[edge height=.45cm]{6}{5}{\textsc{mark}}
\depedge[edge height=1.1cm,edge slant=5,edge style={depblue!90, thick},label style={fill=depblue!35}]{4}{6}{\textsc{xcomp}}
\depedge[edge below,edge height=0.5cm,edge style={depblue!90, thick},label style={fill=depblue!35}]{1}{4}{\textsc{nsubj}}
\depedge[edge below,edge height=1cm,edge style={depgreen!80, thick},label style={fill=depgreen!15}]{1}{6}{\textsc{nsubj:xsubj}}
\end{dependency}
\caption{Control example. The edges of the graph after two passes are in magenta and blue. During the first pass \textcolor{deporange}{\textbf{{weep}}} is stored as it is a dependent of a \texttt{xcomp} relation (highlighted in blue) but it cannot be resolved until \textcolor{depred}{\textbf{{wanted}}} is. \textcolor{depred}{\textbf{{wanted}}} is resolved in the second pass and an \texttt{nsubj} relation (shown in blue) is propagated from the head, \textit{gnomes},  of its conjunct, \textcolor{depred}{\textbf{{quailed}}}. In the third pass this is further propagated to \textcolor{deporange}{\textbf{{weep}}} generating a \texttt{nsubj:xsubj} relation (highlighted in green).}\label{fig:control}
\end{figure}
\subsection{Tuning the rules}
A number of enhancements are relation and language specific and some even lexically conditioned such as control, and not all languages include every enhancement type. 
So the training data is used 
to tune rules to a given language while keeping the rule definitions as generic as possible.

The first type of information needed regards additional lemmas and cases appearing in edges.
For each relation type, the frequency at which \texttt{case} is being added to the relation is obtained.
Similarly for lemma, the algorithm counts the frequency of relation types between a word and its dependent used for lexicalisation since different relations are augmented with different dependents (\texttt{obl} usually uses \texttt{case} where \texttt{acl} prefers \texttt{mark}).
Furthermore, for lemmas, when several dependents have the same relation, it checks 
which is used for lexicalisation.
For \texttt{conj} though, it only checks if there is anything at all since \texttt{conj} is tightly linked to \texttt{cc}.

Each language is tested to see if it is prodrop by comparing the number of root verbs with an overt subject to the number of root verbs without an overt subject.
Whether \texttt{:xsubj} and \texttt{:relsubj} should be added to subjects of controlled predicates and relative clauses is also checked.

The algorithm then checks whether each relation propagates to its governor's conjuncts and under which conditions (the conditions are detailed in Appendix \ref{sec:prop_conj}) and also if it propagates to its own conjuncts.
This is mostly relevant since \texttt{root} usually does not propagate to conjuncts of the main predicate, but in some treebanks it does.

Morphological features are used for detecting relativisers.
For each morphological feature, the number of times it co-occurs with a \texttt{ref} enhanced relation is compared to the number of times it co-occurs with another relation.
While not an arbitrary choice, it is one of the few cases where an enhanced relation does not depend directly on information in the original tree but on information external to the tree, so in theory we could have chosen other clues such as the lemma of the word instead.
These pronouns and adverbs are usually marked with \texttt{PronType=Rel} or \texttt{PronType=Int,Rel}.

Finally, the controlling profile of controlling predicates is learnt.
The system discerns which of the arguments is used as subject of controlled verbs and in which conditions, meaning that we do not count subjects in the absence of other arguments since they become default.

\subsection{Problems}


While our rule-based system performs remarkably well, as can be seen in Table \ref{tab:rule_performance}, with the lowest ELAS being 94.9 on the gold development data, it is challenging to improve across languages simultaneously. 
Besides the expected ambiguity of language, 
there are several issues which limit us, 
some easy to fix, some more complicated, some language specific, and some more general.

\begin{table}[htpb!]\centering\tabcolsep=.001cm\small\begin{tabular}{l    >{\centering\arraybackslash}p{.1em}    >{\centering\arraybackslash}p{3em}    >{\centering\arraybackslash}p{.1em}    >{\centering\arraybackslash}p{3em}    >{\centering\arraybackslash}p{.1em}    >{\centering\arraybackslash}p{3em}     >{\centering\arraybackslash}p{.1em}     >{\centering\arraybackslash}p{3.1em}  }  \toprule  \multicolumn{9}{c}{ELAS} \\\midrule  & & Gold & & Full & & Dist  & & UDPipe \\\textbf{\textit{ Arabic}} &&98.8&&68.4&&67.9&&63.0\\\textbf{\textit{ Bulgarian}} &&98.6&&85.3&&85.2&&81.9\\\textbf{\textit{ Czech}} &&97.9&&82.4&&80.7&&78.4\\\textbf{\textit{ Dutch}} &&98.9&&82.2&&81.0&&73.2\\\textbf{\textit{ English}} &&99.5&&81.2&&79.8&&76.3\\\textbf{\textit{ Estonian}} &&99.2&&80.3&&79.0&&76.7\\\textbf{\textit{ Finnish}} &&97.3&&79.9&&78.0&&73.7\\\textbf{\textit{ French}} &&98.9&&82.3&&82.6&&79.4\\\textbf{\textit{ Italian}} &&99.5&&87.8&&85.9&&84.1\\\textbf{\textit{ Latvian}} &&95.7&&79.3&&78.2&&70.5\\\textbf{\textit{ Lithuanian}} &&98.8&&68.6&&68.9&&60.9\\\textbf{\textit{ Polish}} &&94.9&&78.6&&77.2&&74.7\\\textbf{\textit{ Russian}} &&98.6&&84.4&&82.5&&81.5\\\textbf{\textit{ Slovak}} &&98.8&&77.0&&76.1&&70.5\\\textbf{\textit{ Swedish}} &&98.8&&78.9&&79.0&&73.2\\\textbf{\textit{ Tamil}} &&99.3&&51.2&&55.5&&53.0\\\textbf{\textit{ Ukrainian}} &&95.8&&78.3&&77.5&&72.5\\\textbf{\textit{ Average}} &&98.2&&78.0&&77.3&&73.1\\\bottomrule
\end{tabular}
\caption{Enhanced labelled attachment score for EUD graphs when using gold labelled dependency development treebanks (gold), predicted treebanks with full baseline models (Full), distilled models (Dist), and using UDPipe v2.5 models (UDPipe).}
\label{tab:rule_performance}
\end{table}

\begin{table*}[htpb!]\centering\tabcolsep=.049cm\small\begin{tabular}{l d{3.1}d{3.1}cccc ccccc ccccc}\toprule\multicolumn{1}{c}{}&\multicolumn{1}{c}{Tokens}&\multicolumn{1}{c}{Words}&\multicolumn{1}{c}{Sentences}&\multicolumn{1}{c}{UPOS}&\multicolumn{1}{c}{XPOS}&\multicolumn{1}{c}{UFeats}&\multicolumn{1}{c}{AllTags}&\multicolumn{1}{c}{Lemmas}&\multicolumn{1}{c}{UAS}&\multicolumn{1}{c}{LAS}&\multicolumn{1}{c}{CLAS}&\multicolumn{1}{c}{MLAS}&\multicolumn{1}{c}{BLEX}&\multicolumn{1}{c}{EULAS}&\multicolumn{1}{c}{ELAS}\\\midrule\textbf{\textit{Arabic}}&100.0&94.6&82.1&88.5&84.0&84.2&82.0&88.5&76.5&72.0&68.0&57.0&63.0&70.2&67.8\\\textbf{\textit{Bulgarian}}&99.9&99.9&94.2&97.6&94.3&95.4&93.8&94.6&92.1&88.5&84.5&78.0&77.5&87.3&86.4\\\textbf{\textit{Czech}}&99.9&99.9&93.2&97.8&90.9&90.8&89.7&97.4&88.0&84.1&80.9&70.4&78.6&82.0&79.6\\\textbf{\textit{Dutch}}&99.7&99.7&69.3&92.6&89.9&92.0&89.0&94.4&84.5&80.8&73.7&63.5&68.0&79.3&78.7\\\textbf{\textit{English}}&99.2&99.2&83.8&93.6&92.8&94.1&90.7&95.4&84.8&81.7&77.7&69.0&73.8&80.8&80.1\\\textbf{\textit{Estonian}}&99.7&99.7&90.0&95.0&96.2&92.8&91.0&90.4&82.7&78.2&75.5&67.3&66.2&77.7&76.8\\\textbf{\textit{Finnish}}&99.7&99.7&88.7&94.8&54.5&93.0&51.8&87.1&86.1&82.6&80.0&72.1&67.0&80.8&79.4\\\textbf{\textit{French}}&99.7&99.2&94.3&93.5&99.2&88.8&87.3&94.9&87.8&82.2&74.8&60.5&69.1&81.6&79.5\\\textbf{\textit{Italian}}&99.9&99.8&98.8&97.2&97.0&97.1&96.2&97.4&91.4&89.1&83.8&79.2&80.3&87.6&86.9\\\textbf{\textit{Latvian}}&99.3&99.3&98.7&93.5&84.3&89.5&83.9&92.7&86.0&81.8&78.7&65.9&72.4&79.3&77.8\\\textbf{\textit{Lithuanian}}&99.9&99.9&87.9&90.3&80.7&81.2&79.3&88.8&75.2&69.4&66.0&48.4&56.8&66.6&64.5\\\textbf{\textit{Polish}}&99.4&99.8&97.5&96.4&84.9&83.6&80.3&95.6&90.1&85.9&82.4&62.2&77.8&84.0&77.5\\\textbf{\textit{Russian}}&99.6&99.6&98.8&97.8&99.6&85.3&85.0&96.5&89.3&86.2&83.4&65.5&80.0&84.5&83.3\\\textbf{\textit{Slovak}}&100.0&100.0&85.3&92.9&77.1&80.3&76.7&86.6&85.6&81.5&78.0&56.8&64.8&79.8&76.7\\\textbf{\textit{Swedish}}&99.2&99.2&93.5&93.3&91.0&84.9&83.2&90.0&83.4&79.3&76.0&58.6&67.0&77.9&77.0\\\textbf{\textit{Tamil}}&99.2&94.5&97.5&81.3&76.3&80.5&75.6&84.1&62.5&53.0&48.8&39.9&43.7&53.0&51.7\\\textbf{\textit{Ukrainian}}&99.8&99.8&96.6&94.9&84.0&84.3&83.3&93.6&85.0&81.0&76.4&59.6&70.0&78.4&76.4\\\textbf{\textit{Average}}&99.7&99.1&91.2&93.6&86.9&88.1&83.5&92.2&84.2&79.8&75.8&63.2&69.2&78.3&76.5\\\bottomrule\end{tabular} \caption{Test results evaluated through the official submission site and using our updated distilled model. Our official submission results can be seen in Table \ref{tab:test_results_full_old} in the Appendix.} \label{tab:test_results_full}\end{table*}

On the monolingual front, incomplete, erroneous and inconsistent annotations are the biggest problems.
Incomplete annotation can occur both at the enhanced dependency and the lower level of annotation.
For example in Dutch Alpino, we miss 515 \texttt{ref} relations and thus at least as many enhanced relations from their antecedents, representing two thirds of the missing dependencies.
The bulk of these missed references are relative/interrogative pronouns/adverbs that are not annotated 
with an empty feature column.
We wanted to avoid too many language specific rules and ignored them, leading to more than a thousand missing edges.
Likewise, in some languages not all relativisers (typically interrogative adverbs) are marked as references when they should be according to UD guidelines. 

Erroneous annotations 
can be at lower levels of annotation of the dependency tree, thus when applying rules according to these annotations, erroneous edges are created.
For example in English (EWT), there is the sentence \textit{``Let me know if this is the appropriate steps that you would like to see,''} in which \textit{that} which references \textit{steps} is analysed as the object of \textit{like} (\textit{``you would like the steps to see''} vs. \textit{``you would like to see the steps''}) thus the controlling rule for \textit{like} makes \textit{steps} the subject of \textit{see} in place of \textit{you}.
Annotation errors can also happen in the enhanced structure.
In Russian, for example, a number of nominal modifiers have diverging case information in the feature column and in the enhanced relation one, often \texttt{Case=Gen} with \texttt{nmod:acc}, so the predicted enhanced relation \texttt{nmod:gen} conflicts with the actual annotation.

Latvian offers an example of inconsistent annotation, \texttt{nmod} is extended with either the adposition's lemma or the word's case but never both and the selection of lemma or case for any given word is seemingly arbitrary. 
So it is impossible to devise a rule to address this issue.

However, most of these problems are easily rectified with a system such as ours by checking the agreement of case and lemma information in enhanced relations assuming valid annotation of the underlying data.

On the cross-lingual front, the biggest problem is lack of consistency in annotation conventions. 
Leaving incomplete annotation aside, there are a number of clear divergences. 
The most striking example is the way subjects of passive verbs and more generally enhanced relations are handled in French Sequoia.
These relations receive an extra \texttt{(:)enh} to differentiate them from canonical relations directly taken from the tree, the presence of the column depends mostly on the number of columns in the relation type, if it is a simple relation then a column is used but when it is already a sub-type with a column between the main type and extra information then no column is added.
Not only is this unique to this treebank, 
but it is also redundant since this information can be directly retrieved by looking at the original tree.
There are also a number of more subtle inconsistencies. 
For example, in languages that add lemma information to \texttt{conj} relations, when the coordinating conjunction is a symbol (\& or /), most languages just ignore them and keep the bare \texttt{conj} relation.
However, Swedish uses the special \texttt{conj:sym} relation. 

Beyond these issues, there remain genuine linguistic difficulties.
A difficulty common to all languages is the scope of conjunctions and whether to propagate dependents amongst conjuncts or not.
This is particularly clear with adverbials and obliques that modify verbs.
Due to
their broad semantic range, adverbials can propagate from conjunction heads to dependent conjuncts even if they already have other adverbials, as long as they do not conflict semantically.
Currently in UD, there is no hierarchy amongst dependents of a word, but there could be a form of scope indexing to distinguish a word's direct dependents from dependents of the whole conjunction attached to its head. 

Another difficulty is subject selection in prodrop languages.
Fortunately, the prodrop languages in this shared task have personal and number agreement at least on finite verbs which helps testing the compatibility of the overt subject of a verb with its coordinated verbs or verbs in relative clauses that lack an overt subject.
However, there are prodrop languages that do not mark personal agreement on verbs and do not use relativisers either (e.g. Japanese).
In this case, finding the semantic subject of verbs may  be much more challenging.

\section{Results and Discussion}
Despite focusing on efficiency, our official submission obtained an average ELAS of 74.04 which was the fourth best system (out of 9 full submissions). Our improved score after training distilled models to convergence (or closer to convergence) obtained an average score of 76.14. The full breakdown of these results are shown in Table \ref{tab:test_results_full} and Table \ref{tab:test_results_full_old} in the Appendix. 

Our system is competitive mainly by the grace of our rule-based system which obtains an average 98.20 ELAS when used on the gold development treebanks. And for the most part its performance echoes the quality of the predicted dependencies and tags used by the system as is seen in Table \ref{tab:dev_performance}. 
Having a rule-based system that can perform so well on gold data means that improving the dependency predictions it is based on for a full pipeline will almost always increase ELAS scores. It also means it could be used to generate new data. Although this would be restricted to generating data for pre-existing UD treebanks. 
Furthermore, it could be used to highlight annotation inconsistencies in a given treebank and between different treebanks for the same language.

We also demonstrated that smaller networks can be competitive, even if in this context distillation does not perform as well as previously observed for UD parsing. And beyond that, we show that it is possible to train competitive models with less data and by doing so lowering the energy cost of training parsers. One potentially interesting result is that Tamil performs noticeably better with distillation than either the full baseline model or the small model of the same size trained normally. It has the smallest training treebank out of all the treebanks used in the shared task. The other smaller treebanks also perform better with distillation, e.g the next three smallest treebanks French, Lithuanian, and Swedish all follow this trend but the increase in performance is less pronounced. Perhaps smaller treebanks benefit from what is essentially ensemble training as it tempers a network's penchant for over-fitting.

\section*{Acknowledgments}
This work has received funding from the European Research Council (ERC), under the European Union's Horizon 2020 research and innovation programme (FASTPARSE, grant agreement No 714150), from the  ANSWER-ASAP project (TIN2017-85160-C2-1-R) from MINECO, and from Xunta de Galicia (ED431B 2017/01, ED431G 2019/01). 
\bibliography{acl2020.bib}
\bibliographystyle{acl_natbib}
\appendix
\section{Appendix}\label{appendix:sup}
\subsection{Teacher-student distillation}\label{sec:teacher}
Model distillation is the act of taking one or more models and guiding the training of a single network with these models. It was originally introduced not as a means of creating more efficient models, but as a way of ensemble training with networks \cite{bucilua2006model,ba2014,hinton2015,kuncoro2016}. 

\textit{Teacher-student} distillation, the method used in this work, has been successfully utilised in a number of NLP tasks ranging from machine translation, language modelling, exploring structured linguistic space, and speech recognition \cite{kim2016sequence,lu2017knowledge,liu-etal-2018-distilling,yu2018device}.

In \textit{teacher-student} distillation, the \textit{teacher} guides the training of another model, the \textit{student}, which in our experiments is smaller. The \textit{student} explicitly uses the information of the larger model by comparing the probability distribution of the respective model's output layer. We use the Kullback-Leibler divergence to obtain the loss between these two distributions:
\begin{equation}
    \mathcal{L}_{KL}= -\sum_{t \in b}\sum_{i} P(\textbf{x}_i)\log\frac{P(\textbf{x}_i)}{Q(\textbf{x}_i)}
\end{equation}
where $P$ is the probability distribution from the teacher's softmax layer, $Q$ is the probability distribution from the student's, and $\textbf{x}_i$ is input vector to the softmax corresponding to token $w_i$ of a given tree $t$ for all trees in batch $b$. 

For our implementation there are two probability distributions as we are using a Biaffine parser, one for head predictions and one for label predictions.\footnote{The PyTorch implementation used can be found at:\\ \url{www.github.com/zysite/biaffine-parser}} 

The \textit{student} is also trained directly on the gold heads and labels using a categorical cross entropy loss, e.g. for the loss on head predictions:
\begin{equation}
    \mathcal{L}_{CE}= -\sum_{t \in b}\sum_{i} \log p(h_i|\textbf{x}_i)
\end{equation}
where $h_i$ is the true head position for token $w_{i}$, corresponding to the softmax layer input vector $\textbf{x}_i$, of tree $t$ in batch $b$. 

The total loss is therefore the combination of the Kullback-Leibler loss between the probability distributions of the \textit{teacher} and the \textit{student} for both head and label predictions with the cross entropy loss between the \textit{student} predictions and the gold data:
\begin{align}
    \mathcal{L} = \mathcal{L}_{KL}&(T_{h}, S_{h}) + \mathcal{L}_{KL}(T_{lab}, S_{lab})\nonumber\\ &+ \mathcal{L}_{CE}(h) + \mathcal{L}_{CE}(lab)
\end{align}
where $\mathcal{L}_{CE}(h)$ is the loss for the student's predicted head positions, $\mathcal{L}_{CE}(lab)$ is the loss for the student's predicted arc label, $\mathcal{L}_{KL}(T_h, S_h)$ is the loss between the teacher's probability distribution for arc predictions and that of the student, and $\mathcal{L}_{KL}(T_{lab}, S_{lab})$ is the loss between label distributions.
\begin{figure}[t]
    \centering
    \includegraphics[width=0.99\linewidth]{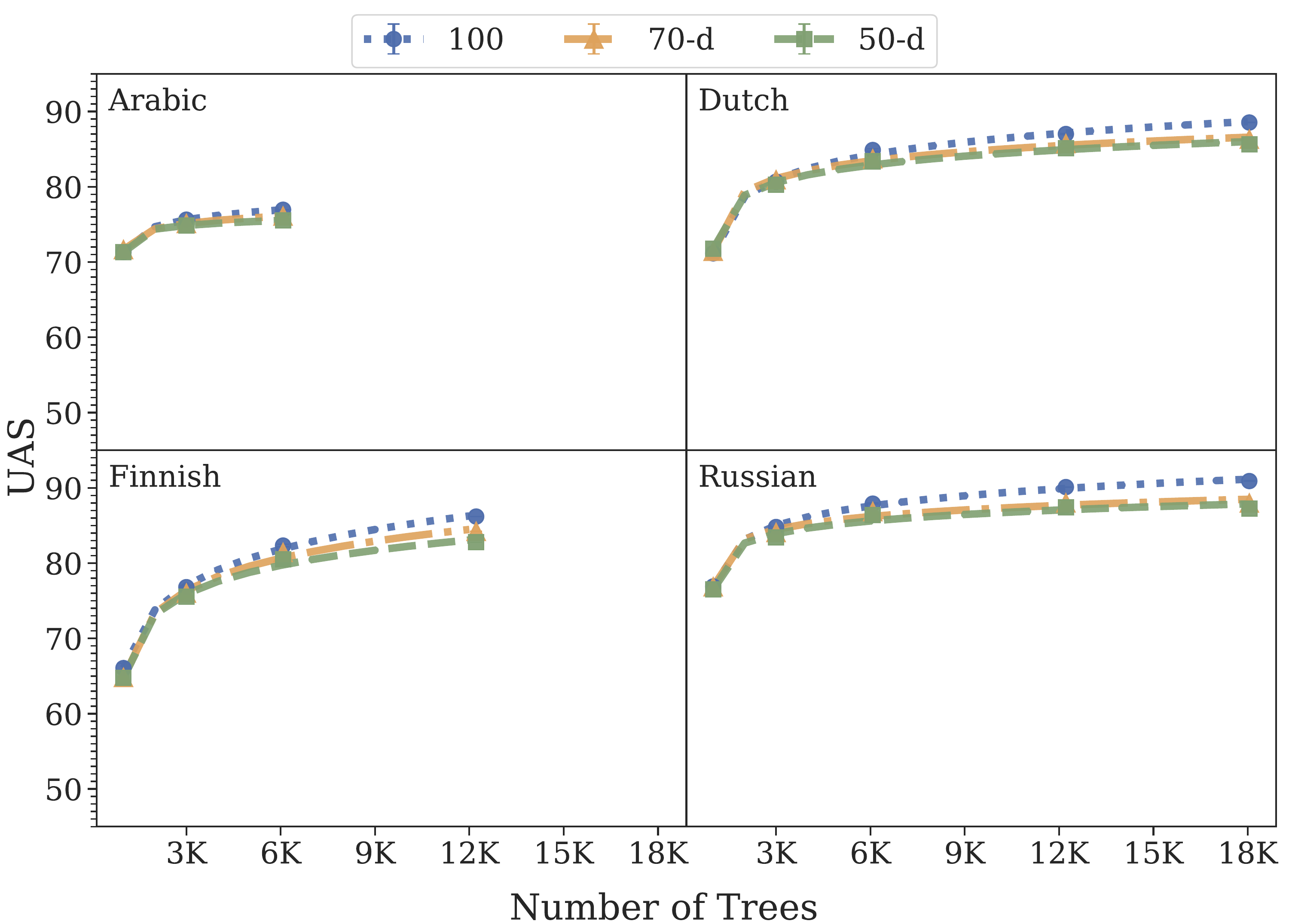}
    \caption{UAS for different models for Arabic, Dutch, Finnish, and Russian development treebanks.}
    \label{fig:uas}
\end{figure}

\begin{table}[htpb!]
\footnotesize
    \centering
        \tabcolsep=.25cm  
    \begin{tabular}{l p{0.5em} r}
    \toprule
    \textbf{hyperparameter} & & \textbf{value}\\
    \toprule
         word embedding dimensions& & 100\\
         char embedding dimensions&  & 32\\
         char BiLSTM dimensions& & 100\\
         embedding dropout&  & 0.33 \\
         BiLSTM dimensions&  & 400 (200)  \\
         BiLSTM layers&  & 3 \\
         arc MLP dimensions&  & 500 (250)\\
         label MLP dimensions&  & 100 (50)\\
         MLP layers&  & 1 \\
         learning rate&  & 0.2 \\
         dropout&  & 0.33 \\
         momentum&  & 0.9 \\
         L2 norm $\lambda$&  & 0.9\\
         annealing&  & 0.75$^{\wedge}(\nicefrac{t}{5000})$\\
         $\epsilon$& & 1$\times 10^{-12}$\\
         optimiser&  & Adam \\
         loss function&  & cross entropy \\
         epochs&  & 100 \\
         min vocab freq. & &2 \\
    \bottomrule
    \end{tabular}
        
    \caption{Hyperparameters for baseline models. The values in parentheses show the values for the distilled and small models used in the main analysis of the shared task.}
    \label{tab:experimental_hyperparameters}
\end{table}
\begin{table*}[htpb!]
  \centering
  \tabcolsep=.07cm
  \small
  \begin{tabular}{l
      >{\raggedleft\arraybackslash}p{3em}
      >{\raggedleft\arraybackslash}p{2.5em}
      >{\centering\arraybackslash}p{2.5em}
      >{\centering\arraybackslash}p{2.em}
      >{\raggedleft\arraybackslash}p{0.5em}
      >{\raggedleft\arraybackslash}p{3em}
      >{\raggedleft\arraybackslash}p{2.5em}
      >{\centering\arraybackslash}p{2.5em}
      >{\centering\arraybackslash}p{2.em}
      >{\raggedleft\arraybackslash}p{0.5em}
      >{\centering\arraybackslash}p{2.5em}
    }
    \toprule
    & \multicolumn{4}{c}{original} & & \multicolumn{4}{c}{sample} & &\\\cmidrule{2-5}\cmidrule{7-10}
    & \multicolumn{1}{c}{\hspace{0.5em}trees} & \multicolumn{1}{c}{\hspace{0.75em}mL} & \multicolumn{1}{c}{mDD} & \multicolumn{1}{c}{NP\%}  &  &  \multicolumn{1}{c}{\hspace{0.5em}trees} & \multicolumn{1}{c}{\hspace{0.75em}mL} & \multicolumn{1}{c}{mDD} & \multicolumn{1}{c}{NP\%} & & \multicolumn{1}{c}{2s-KS$_\textrm{L}$}\\
    \toprule
\multirow{1}{*}{\textit{\textbf{Czech}}} & & & & & \\\hspace{1em}-CAC & 23478&20.1&3.7&2.5& &3016&19.9&3.7&2.5& &0.014\\\hspace{1em}-FicTree & 10160&13.2&3.6&3.8& &1305&13.1&3.6&3.8& &0.016\\\hspace{1em}-PDT & 68495&17.1&3.7&2.7& &8800&17.1&3.7&2.6& &0.007\\\hspace{1em}-combined & 102133&17.4&3.7&2.7& &13121&17.4&3.7&2.7& &0.007\\\multirow{1}{*}{\textit{\textbf{Dutch}}} & & & & & \\\hspace{1em}-Alpino & 12264&15.2&4.0&4.5& &8915&15.2&4.0&4.4& &0.004\\\hspace{1em}-LassySmall & 5787&13.0&3.7&2.0& &4206&13.0&3.7&1.9& &0.007\\\hspace{1em}-combined & 18051&14.5&3.9&3.8& &13121&14.5&3.9&3.7& &0.004\\\multirow{1}{*}{\textit{\textbf{Estonian}}} & & & & & \\\hspace{1em}-EDT & 24633&14.0&3.6&0.8& &12552&14.1&3.6&0.8& &0.010\\\hspace{1em}-EWT & 1116&15.4&3.8&1.5& &569&15.4&3.8&1.6& &0.027\\\hspace{1em}-combined & 25749&14.1&3.6&0.9& &13121&14.1&3.6&0.8& &0.009\\\multirow{1}{*}{\textit{\textbf{Polish}}} & & & & & \\\hspace{1em}-LFG & 13774&7.6&2.8&0.3& &5738&7.6&2.8&0.3& &0.006\\\hspace{1em}-PDB & 17722&15.9&3.4&1.4& &7383&15.9&3.4&1.5& &0.008\\\hspace{1em}-combined & 31496&12.3&3.3&1.1& &13121&12.3&3.3&1.2& &0.003\\\multirow{1}{*}{\textit{\textbf{Russian}}} & & & & & \\\hspace{1em}-SynTagRus & 48814&17.8&3.6&1.6& &13121&17.8&3.6&1.6& &0.004\\\bottomrule
\end{tabular}
 \caption{Analysis of renormalised treebank samples: 2s-KS is the two-sample Kolmogorov-Smirnov test comparing the sentence-length distributions of the original and the sample treebanks (where values close to 0 suggest samples are not from different distributions, and values approaching 1 suggest otherwise); trees is the number of trees; mL is the mean sentence length; mDD the mean dependency distance; and NP\% is the percentage of non-projective arcs. Where we use the combined sample (or just the sample for Russian-SynTagRus) for training.}
 \label{tab:treebank_samples}
\end{table*}

\begin{table}[htpb!]
\centering
\tabcolsep=.049cm
\small
\begin{tabular}{rcccrcc}
\toprule
&\multicolumn{1}{c}{UAS} & LAS  & \phantom{} & & \multicolumn{1}{c}{UAS} & LAS      \\\midrule
\multirow{1}{*}{\textit{\textbf{Arabic}}} & & & & \multirow{1}{*}{\textit{\textbf{Bulgarian}}} \\small & \multicolumn{1}{c}{76.9} & 72.5 &  & small& \multicolumn{1}{c}{91.6} & 87.6\\dist & \multicolumn{1}{c}{76.5} & 72.3 &  & dist& \multicolumn{1}{c}{91.6} & 87.6 \\
\multirow{1}{*}{\textit{\textbf{Czech}}} & & & & \multirow{1}{*}{\textit{\textbf{Dutch}}} \\small & \multicolumn{1}{c}{89.5} & 86.0 &  & small& \multicolumn{1}{c}{87.2} & 83.3\\dist & \multicolumn{1}{c}{89.0} & 85.3 &  & dist& \multicolumn{1}{c}{86.7} & 82.9 \\
\multirow{1}{*}{\textit{\textbf{English}}} & & & & \multirow{1}{*}{\textit{\textbf{Estonian}}} \\small & \multicolumn{1}{c}{85.0} & 81.9 &  & small& \multicolumn{1}{c}{85.2} & 80.9\\dist & \multicolumn{1}{c}{84.4} & 81.2 &  & dist& \multicolumn{1}{c}{84.7} & 80.2 \\
\multirow{1}{*}{\textit{\textbf{Finnish}}} & & & & \multirow{1}{*}{\textit{\textbf{French}}} \\small & \multicolumn{1}{c}{85.8} & 82.2 &  & small& \multicolumn{1}{c}{88.1} & 85.5\\dist & \multicolumn{1}{c}{85.1} & 81.3 &  & dist& \multicolumn{1}{c}{88.5} & 85.8 \\
\multirow{1}{*}{\textit{\textbf{Italian}}} & & & & \multirow{1}{*}{\textit{\textbf{Latvian}}} \\small & \multicolumn{1}{c}{91.3} & 89.0 &  & small& \multicolumn{1}{c}{86.3} & 82.4\\dist & \multicolumn{1}{c}{90.3} & 87.8 &  & dist& \multicolumn{1}{c}{86.0} & 81.9 \\
\multirow{1}{*}{\textit{\textbf{Lithuanian}}} & & & & \multirow{1}{*}{\textit{\textbf{Polish}}} \\small & \multicolumn{1}{c}{76.7} & 71.5 &  & small& \multicolumn{1}{c}{90.5} & 86.4\\dist & \multicolumn{1}{c}{78.0} & 73.0 &  & dist& \multicolumn{1}{c}{90.2} & 86.0 \\
\multirow{1}{*}{\textit{\textbf{Russian}}} & & & & \multirow{1}{*}{\textit{\textbf{Slovak}}} \\small & \multicolumn{1}{c}{89.5} & 86.3 &  & small& \multicolumn{1}{c}{85.6} & 81.7\\dist & \multicolumn{1}{c}{88.9} & 85.5 &  & dist& \multicolumn{1}{c}{84.7} & 80.7 \\
\multirow{1}{*}{\textit{\textbf{Swedish}}} & & & & \multirow{1}{*}{\textit{\textbf{Tamil}}} \\small & \multicolumn{1}{c}{84.5} & 80.8 &  & small& \multicolumn{1}{c}{63.7} & 55.7\\dist & \multicolumn{1}{c}{85.3} & 81.6 &  & dist& \multicolumn{1}{c}{64.0} & 56.9 \\
\multirow{1}{*}{\textit{\textbf{Ukrainian}}} & & & & \multirow{1}{*}{\textit{\textbf{Average}}} \\small & \multicolumn{1}{c}{86.8} & 82.6 &  & small& \multicolumn{1}{c}{85.0} & 80.9\\dist & \multicolumn{1}{c}{86.6} & 82.5 &  & dist& \multicolumn{1}{c}{84.7} & 80.7 \\\bottomrule
\end{tabular}
\caption{Comparison of attachment scores for the development treebanks for distilled (dist) models and models with the same parameters (small) trained normally.}
\label{tab:small_performance}
\end{table}

\begin{table*}[htpb!]\centering\tabcolsep=.044cm\small\begin{tabular}{l d{3.1}d{3.1}cccc ccccc ccccc}
\toprule&\multicolumn{1}{c}{Tokens}&\multicolumn{1}{c}{Words}&Sentences&UPOS&XPOS&UFeats&AllTags&Lemmas&UAS&LAS&CLAS&MLAS&BLEX&EULAS&ELAS\\\midrule\textbf{\textit{Arabic}}&100.0&94.6&82.1&88.5&84.0&84.2&82.0&88.5&75.8&71.2&66.8&56.1&62.0&69.2&66.9\\\textbf{\textit{Bulgarian}}&99.9&99.9&94.2&97.6&94.3&95.4&93.8&94.6&91.1&87.0&82.3&75.8&75.5&85.8&84.9\\\textbf{\textit{Czech}}&99.9&99.9&93.2&97.8&90.9&90.8&89.7&97.4&86.2&81.8&78.1&67.9&75.8&79.6&77.2\\\textbf{\textit{Dutch}}&99.7&99.7&69.3&92.6&89.9&92.0&89.0&94.4&83.4&79.4&71.9&61.9&66.4&78.0&77.4\\\textbf{\textit{English}}&99.2&99.2&83.8&93.6&92.8&94.1&90.7&95.4&83.7&80.1&75.8&67.2&72.1&79.2&78.5\\\textbf{\textit{Estonian}}&99.7&99.7&90.0&95.0&96.2&92.8&91.0&90.4&80.7&75.5&72.7&64.6&63.8&75.0&74.1\\\textbf{\textit{Finnish}}&99.7&99.7&88.7&94.8&54.5&93.0&51.8&87.1&84.1&79.7&76.5&69.0&64.3&77.8&75.7\\\textbf{\textit{French}}&99.7&99.2&94.3&93.5&99.2&88.8&87.3&94.9&87.2&80.6&72.1&58.3&66.7&80.1&77.8\\\textbf{\textit{Italian}}&99.9&99.8&98.8&97.2&97.0&97.1&96.2&97.4&90.2&87.4&81.4&76.8&77.9&85.9&84.8\\\textbf{\textit{Latvian}}&99.3&99.3&98.7&93.5&84.3&89.5&83.9&92.7&84.4&79.7&76.1&63.6&70.0&77.2&75.6\\\textbf{\textit{Lithuanian}}&99.9&99.9&87.9&90.3&80.7&81.2&79.3&88.8&72.9&66.3&62.6&45.9&54.3&63.7&61.4\\\textbf{\textit{Polish}}&99.4&99.8&97.5&96.4&84.9&83.6&80.3&95.6&88.4&83.4&79.4&60.1&75.0&81.4&74.5\\\textbf{\textit{Russian}}&99.6&99.6&98.8&97.8&99.6&85.3&85.0&96.5&86.8&83.2&80.0&62.8&76.7&81.7&80.3\\\textbf{\textit{Slovak}}&100.0&100.0&85.3&92.9&77.1&80.3&76.7&86.6&83.2&78.3&73.9&53.8&61.6&76.5&73.5\\\textbf{\textit{Swedish}}&99.2&99.2&93.5&93.3&91.0&84.9&83.2&90.0&82.2&77.6&73.7&56.8&64.9&76.2&75.2\\\textbf{\textit{Tamil}}&99.2&94.5&97.5&81.3&76.3&80.5&75.6&84.1&59.6&48.8&43.6&35.5&39.6&48.1&47.0\\\textbf{\textit{Ukrainian}}&99.8&99.8&96.6&94.9&84.0&84.3&83.3&93.6&83.4&78.7&73.6&57.8&67.4&76.2&74.0\\\textbf{\textit{Average}}&99.7&99.1&91.2&93.6&86.9&88.1&83.5&92.2&82.5&77.6&73.0&60.8&66.7&76.0&74.0\\\bottomrule\end{tabular} 
\caption{Full test results for our official submission using the shared task's submission site for evaluation.}\label{tab:test_results_full_old}
\end{table*}

\subsection{Details of dependency enhancements}\label{sec:enh_dets}

In this section we give more details about the enhancement of dependency relations and about the processing subtleties of relative clauses, controlled predicates, and conjunctions.

Most of the original dependencies are kept in the enhanced structure, but they can undergo a number of cosmetic changes.
In the simplest case, the relation type $t$ is just appended to the index $h$ of the word's governor to give the relation $h:t$.
Sometimes, during the process the relation type is slightly modified. 
In Estonian (EDT and EWT) some complex relations such as \texttt{compound:prt} or \texttt{csubj:cop} are truncated and only the first part is kept.
Conversely, in French (Sequoia) some relations receive extra information, such as subjects of passives \texttt{nsubj:pass} that are augmented with \texttt{xoxobj} stating they are the semantic object of their head.

Some relations receive extra lexical and morphological information.
Conjuncts marked with \texttt{conj} usually receive the lemma of the coordinating conjunction (\texttt{cc}).
Likewise, adverbial and adjectival clauses (\texttt{advcl} and \texttt{acl}) receive the lemma of the word (\texttt{mark}) that introduces them.
Nominal modifiers and obliques (\texttt{nmod} and \texttt{obl}) can receive the lemma of the adposition that introduces them (often marked with the \texttt{case} relation).
Furthermore \texttt{nmod} and \texttt{obl} can also receive case information about the word itself.
When the introducing marker is not a word but a fixed expression such as \textit{``as well as''} then the \textit{long lemma} composed of the lemmas of each word in the expression (marked by the \texttt{fixed} relation) is used, for example \texttt{conj:as\_well\_as}.

\paragraph{Relative clauses}
The only relations from the original tree that are not kept in the enhanced structure are those whose dependent is an anaphoric pronoun or adverb used to introduce a relative clause.
Instead, the dependent (pronoun or adverb) is linked to its antecedent by an edge labelled \texttt{ref}.
A new edge is then added between the original head of the reference and its antecedent of the same type as the original relation in order to show the argument structure of the clause.
Thus, relative clauses are the first phenomenon that creates edges that are not present in the original tree.
Their structure is however relatively simple since they can at most create one extra edge and replace one.

There are nonetheless two subtleties with relative clauses.
First, in some languages, such as English, relative pronouns are not necessary.
In these cases, while there are restrictions on the role the antecedent can fill, we need to infer its actual role from the sentence.
Second, there may be several words that look like relativisers in a relative clause even outside conjunction.
Often, only one of them is a leaf node, the others introducing further embedded clauses. 
Only in Finnish (TDT) did we find instances of multiple relative pronouns attaching to the same verb and each being marked as the reference of another word in the sentence.

\paragraph{Control}
A second phenomenon that creates new dependencies is control, where the subject of an embedded clause is not overt and is provided by one of its governor's arguments.
For example in the English sentence \textit{``I want you to go,''} the semantic subject of the verb \textit{go} is the object of the main verb, namely \textit{you}.
In such a case, an additional relation is added to the structure to represent the dependency of the word \textit{you} to the embedded predicate \textit{go}.
These structures are marked by a \texttt{xcomp} relation between the embedded predicate and its governor in the original tree.
The identity of the new subject depends usually on the governing predicate and its argument structure.
So it is mostly a matter of knowing the governing profile of each lexical item given their argument structure.
For example, the subject of a predicate embedded in a \textit{want to} clause is the object of the \textit{want to} clause if present, its subject otherwise.
Control is also quite simple since it has a limited span.

\paragraph{Conjunction}
 The vast majority of new edges are created by conjunctions and is much harder to handle than the two previous phenomena.
Contrary to relative clauses and control, conjunction has no direction in the sense that it can occur both at the governor level and at the dependent level.
In \textit{``Mary and Sam bought strawberries,''} the conjunction \textit{``Mary and Sam''} occurs at the dependent level and both \textit{Mary} and \textit{Sam} are subject of the verb \textit{bought}.
In \textit{``Mary bought strawberries and ate them,''} the conjunction is now at the governor level and \textit{Mary} is the subject of both \textit{bought} and \textit{ate}.
So unlike relative clauses where 
one merely needs to find the relativiser's antecedent higher up in the tree, or control where one needs to look for the controlled subject amongst the arguments of the controlling predicate, conjunctions can have repercussions both higher up and lower down in the structure at the same time.

The easiest case for conjunction is when it occurs at the dependent level. One just needs to propagate the relation existing between the head of the conjunction and its governor to the other conjuncts.
In the case of conjunction at the governor level, things are more complicated.
While dependents don't tend to propagate up a conjunction chain but only down, they can be blocked by a number of reasons.
For example in \textit{``Mary bought and ate strawberries,''} the object \textit{strawberries} should attach to \textit{bought} in the tree and only propagate down to \textit{ate}.
But in \textit{``Mary spoke and ate strawberries,''} \textit{strawberries} should attach to \textit{ate} and not propagate up to \textit{spoke}, even though \textit{speak} can also have direct objects.
And in \textit{``Mary bought strawberries and ate,''} \textit{strawberries} does not propagate down to \textit{ate} since it appears before it in the sentence.
However, the conditions under which certain dependents do or do not propagate to their governor's conjuncts are both language and relation specific.
In a given language, objects need not behave like subjects nor like determiners or adverbials.
Often if a relation slot (object, subject, determiner) is already filled for a given word, it will block the propagation of the same relation from higher up in the conjunction chain, but it need not always be the case, especially with adverbials.
But even an empty slot does not always guarantee propagation, especially in case marking and prodrop languages where morphological consideration play a major role as well.
So we need to learn the propagation conditions for each relation type on a per language basis.

In our system, we keep track of dependents of \texttt{conj} relations during the first traversal of a sentence and handle them in the second pass.
The main reason for not processing conjuncts as soon as they arrive in the sentence is that some of their dependents (objects, adjectives or adverbials) can appear later and thus would require extra processing.
For example, in \textit{``Mary bought and ate strawberries,''} the object of both verbs only appears after the conjunct \textit{ate}, so upon first seeing \textit{ate}, \textit{bought} does not have any object to be propagated.

\subsubsection{Conjunction propagating conditions}\label{sec:prop_conj}

We use two sets of conditions in order to guide the propagation of dependents to their governor's conjuncts.
The first is about relation types already attached to these conjuncts.
Usually an object or a subject does not attach to a verb that already has these slots filled.
So for each relation, we measure three frequencies.
The frequency at which it co-occurs with other types under its main governor (in the tree), the frequency at which it co-occurs with other types under its conjunct governors (in the enhanced structure) and the frequency at which it does not co-occur with other types because it does not propagate to its governor's conjunct.
Any relation with which it co-occurs under its main governor cannot be blocking propagation.
Then if a relation is more often than not associated with conjunct governors to which the current relation did not propagate, it is considered a blocking relation.
In practice this means that a \texttt{subj} does not propagate to a conjunct of its governor that already has an \texttt{expl}, for example.

The second condition is based on matching morphological information.
For every relation and morphological category (tense, case, aspect, and so on), we measure how often the value of a category agrees or disagrees between the governor and its conjunct (of the same UPOS tag) when the relation propagates and when it does not.
If a category disagrees more often than not between conjuncts which the relation did not propagate, then we assume that the category needs to agree for that relation.


\subsection{Curious quibbles and questionable jiggery-pokery}\label{sec:future}
While being above 94.9 ELAS for all languages, our rule-based system could still be improved to better capture enhanced structures.
There are three main points for further improvement.

Upon reviewing the code for the rule-based system, 
we realised that we catch arguments of relative clauses only in presence of a relativiser that receives the \texttt{ref} relation.
This means that we miss a number of relations involved in relative clauses.
It remained unnoticed because of all the languages in the shared task, most use relative pronouns/adverbs to introduce relative clauses.
In fact the only language that does not have relative pronouns, Tamil, is not yet annotated with relative clauses and it might not even be relevant. 
Our methodology here is to look for an antecedent when we have a relative pronoun, but we could do the opposite and look for potential relative pronouns when we have a relative clause.
The latter should indeed be more language agnostic and work even when there are no relativisers involved.

A second point of improvement has to do with subject finding in controlled predicates.
In our current system, the controlling behaviour of each controlling construction is gathered from the training data, and if we encounter an out of vocabulary construction at prediction time the subject is used by default.
But further consideration 
showed that the object might be a more sensible default option when available.
It would, however, be more interesting to learn the default behaviour on a per language basis. 

Thirdly, due to the march of time, we hard-coded a number of heuristic thresholds used to fine-tune the system.
For example, to see if a language is prodrop, we compare the number of root verbs with overt subjects with the number of root verbs without a subject.
If at least a third of root verbs do not have an overt subject then that language was considered prodrop.
This is clearly not satisfying since this ratio can greatly vary from language to language and from genre to genre.
Furthermore, some languages may not be generally prodrop, but ommit syntactic subjects in impersonal constructions, such as Hebrew, or be prodrop only for certain tenses. 

\end{document}